\documentclass[letterpaper]{article} 
\usepackage[preprint]{aaai2027}  
\usepackage[hyphens]{url}  
\usepackage{graphicx} 
\usepackage{natbib}  
\usepackage{caption} 
\usepackage{algorithm}
\usepackage{algorithmic}

\usepackage{newfloat}
\usepackage{listings}
\DeclareCaptionStyle{ruled}{labelfont=normalfont,labelsep=colon,strut=off} 
\floatstyle{ruled}
\newfloat{listing}{tb}{lst}{}
\floatname{listing}{Listing}

\usepackage{booktabs}

\usepackage{amssymb}

\title{Supervising the Path to Fine Scales: GalerkinFlow for Scientific-Field and Image Super-Resolution}
\author{
    Zikang Zhan
}
\affiliations{
    \textsuperscript{\rm }University College London\\

    London, United Kingdom\\
    zikang.zhan.22@ucl.ac.uk
}

\begin{document}

\maketitle

\begin{abstract}
Most super-resolution models learn from paired data by supervising only the final high-resolution output. This provides little control over how the prediction should evolve between the downsampled observation and its fine target. We introduce GalerkinFlow, an equation-agnostic framework that turns each coarse--fine pair into supervision along an entire reconstruction path. At a random sample of intermediate states on the reconstruction path, the model predicts the coarse-to-fine residual velocity and uses coarse-anchor point to define a pseudo-endpoint. We show that the reconstruction loss of this pseudo-endpoint is exactly related to the intermediate velocity loss through a known time-dependent weight. Consequently, every intermediate state contributes supervision toward the same fine target, rather than serving only as an internal step toward an endpoint loss. Because intermediate states already reveal part of the missing fine-scale structure, we additionally supervise the coarse endpoint used during one-step inference. A finite-difference objective further constrains local spatial variation. GalerkinFlow combines convolutional features with scale-conditioned Galerkin operator mixing and requires no governing equation or physical metadata. It achieves the lowest raw-space errors among the evaluated equation-agnostic baselines on Navier--Stokes and Darcy Flow, while remaining competitive on DIV2K.
\end{abstract}


\section{Introduction}

Super-resolution (SR) is typically framed as a problem of architectural design and model expressiveness. While image-based methods focus on developing stronger local feature extractors, implicit representations, or generative priors, scientific methods often incorporate operator layers, spectral structures, or knowledge of the underlying dynamics. Although these directions are valuable, they leave a fundamental question relatively underexplored: how much supervision can actually be extracted from a single paired coarse-and-fine sample?

We investigate this question within scientific domains where governing equations and physical metadata are unavailable. In our setting, the model receives only a downsampled field, its paired fine target during training, and a requested scale. It is not provided with differential operators, coefficients, forcing terms, boundary conditions, temporal derivatives, or PDE residuals. While this equation-agnostic setting is more restricted than physics-informed reconstruction, it is broadly applicable: the identical training procedure can be utilized whether the data is simulated, empirically measured, or completely decoupled from its original solver.

Most deterministic methods treat a coarse-fine pair merely as a single endpoint constraint. After aligning the coarse observation onto the target grid, they learn the maximum corrective mapping directly from the coarse anchor to the fine target. While this trains the model to match the exact state encountered during inference, it fails to capture how a progressively refined field should behave when part of the missing structure is already present. Numerous endpoint mappings can perfectly fit the observed pairs without establishing a consistent trajectory across partially restored fields. Fundamentally, a direct decoder lacks an explicit restoration state or progress variable upon which a sequential constraint could act.

Our foundational observation is that a paired sample defines strictly more than its two endpoints. The linear trajectory between the coarse anchor and the fine target yields a continuum of partially restored states, and the pair determines the exact residual from each such state back to the fine endpoint. Consequently, GalerkinFlow samples these intermediate states and learns a progress-conditioned residual field. This shifts the training paradigm from enforcing a single terminal constraint per pair to imposing state-residual constraints along a task-relevant trajectory. This formulation does not produce new information; rather, it fully exploits the geometric relationship already embedded within the pair. Unlike noise-to-data generative flows, our source is the deterministic coarse observation, rendering the entire restoration path strictly deterministic.

Nevertheless, intermediate states must be handled cautiously. Except at the strict coarse endpoint ($t=0$), any constructed intermediate state inherently leaks partial information regarding the fine displacement. A model conditioned on both this intermediate state and the coarse anchor could exploit this partial target information as a shortcut. In contrast, one-step inference during testing begins solely at the coarse endpoint and lacks this advantage. To mitigate this training-inference discrepancy, we retain an explicit coarse-anchor reconstruction objective. Path and endpoint supervision thus play complementary roles: the former constrains the residual field across the restoration progress, while the latter anchors the actual inverse problem at $t=0$. Additionally, a finite-difference loss is applied to constrain spatial variations at the reconstructed endpoint. Notably, none of these terms requires evaluating a governing PDE.

We parameterize the residual field using an established local-global architecture. A Residual Dense Network (RDN) encoder \cite{zhang2018rdn} extracts neighborhood-aware features from the coarse anchor, while Galerkin attention \cite{cao2021galerkin} provides domain-wide mixing using parameters independent of the target grid's spatial resolution. Scale and target-cell mappings enable this unified model definition to operate across various uniform grids. We do not claim the CNN-Galerkin combination itself as a novelty; SRNO \cite{wei2023srno} has successfully applied convolutional encoders with Galerkin-type operator layers for image SR. Instead, our core contribution lies in the coarse-anchored path objective built around this backbone.

Our primary empirical claims center on scientific reconstruction. On Navier-Stokes and Darcy Flow datasets, GalerkinFlow improves upon all evaluated equation-agnostic baselines (including U-Net, FNO, U-NO, and SRNO) across all reported raw-space error metrics at both $2\times$ and $4\times$ scales. Furthermore, we conduct RGB experiments on DIV2K and four transfer datasets as a cross-domain stress test. In these settings, GalerkinFlow obtains strong PSNR and SSIM, while LPIPS reveals a more mixed perceptual comparison with image-specialized baselines.

Our contributions are:
\begin{itemize}
\item We formulate each coarse-fine pair as a deterministic path, where intermediate states carry exact endpoint-residual targets. We explicitly relate residual-field error along this restorative path to the one-step endpoint error derived from each sampled state.
\item We introduce a coarse-anchored Galerkin residual-field model that combines pathwise supervision with an explicit $t=0$ reconstruction objective. The latter prevents training from relying only on fine structure already exposed by intermediate states.
\item We demonstrate comprehensive evaluations across two PDE-field benchmarks and natural RGB images. GalerkinFlow yields substantial improvements in raw-space PDE accuracy and strong RGB fidelity scores. Concurrently, the LPIPS results delimit the scope of our claims: perceptual-feature alignment is not uniformly improved by the current objective and backbone.
\end{itemize}

\section{Related Work}

\subsection{Endpoint Super-Resolution}

Most supervised SR methods learn a direct map from a coarse input to a fine endpoint. U-Net-style encoder--decoders \cite{ronneberger2015unet} and RDN residual dense blocks \cite{zhang2018rdn} construct robust local representations. LIIF instead queries local latent codes and coordinates, decoupling the output grid from a fixed pixel-shuffle head \cite{chen2021liif}. SwinIR introduces shifted-window transformer blocks to image restoration, remaining a strong image-SR baseline for image SR \cite{liang2021swinir}. Fundamentally, these methods rely exclusively on paired endpoints for training. In scientific data, convolutional networks have similarly reconstructed turbulent fields without placing a governing equation in the network or loss function \cite{fukami2019superresolution}. While such methods are equation-agnostic, their standard objectives do not specify how the predictor should behave on partially restored intermediate fields.

Neural operators learn mappings between function spaces and can reuse parameters across compatible discretizations \cite{kovachki2023neuraloperator}. FNO performs global kernel operations in Fourier space \cite{li2021fno}, while U-NO uses a U-shaped multiresolution operator to increase depth with controlled memory \cite{rahman2023uno}. SRNO combines an image encoder with Galerkin-type attention for continuous image SR \cite{wei2023srno}. More recently, HiNOTE develops a hierarchical Galerkin operator and a frequency-aware loss prior for arbitrary-scale scientific SR \cite{luo2024hinote}. These works motivate resolution-compatible and global architectures. GalerkinFlow adopts related components but changes the supervised object from a terminal field to a progress-conditioned endpoint residual field.

\subsection{Trajectory-Based Super-Resolution}

Flow matching fits vector fields at sampled states along prescribed probability paths \cite{lipman2023flowmatching}, while rectified flow studies straight transport and efficient integration \citep{liu2023rectified}. Functional Flow Matching (FFM) extends this construction to function spaces \citep{kerrigan2024ffm}. Conditional generative models have also represented the multiple fine outputs compatible with a coarse input. SRFlow learns a conditional image distribution with a normalizing flow \citep{lugmayr2020srflow}, and PSRFlow uses a flow-based latent distribution to quantify uncertainty in scientific SR \citep{shen2023psrflow}. Adaptive Flow Matching separates deterministic large scales from stochastic small-scale content for weather and Kolmogorov-flow downscaling \citep{fotiadis2025adaptive}.

Recent image methods bring trajectory and endpoint constraints closer together. FlowSR combines rectified flow, consistency learning, and explicit high-resolution regularization for one-step image SR \citep{xu2025flowsr}. CTMSR maps points on a probability-flow trajectory to a common output and adds distribution trajectory matching \citep{you2025ctmsr}. RFMSR starts from a low-quality-centered latent distribution and retains velocity supervision while adding end-to-end single-step training \citep{huang2026rfmsr}. These works show that path and endpoint supervision have already been successfully combined for image SR tasks.

In contrast, our objective is deterministic and conceptually more constrained. Unlike generative frameworks, our source is strictly the observed coarse field rather than Gaussian noise, and we do not employ a teacher model or a distribution-matching loss. The supervised vector-field target is the paired endpoint residual from each sampled intermediate state. The main experiments use one Euler step from the lifted coarse anchor. During training, intermediate states define additional residual targets along the restoration path, and $t$ strictly denotes restoration progress.

\subsection{Derivative-Aware Supervision}

Pointwise values and spatial derivatives describe different aspects of a field. Sobolev training explores objectives that match function values together with derivatives \citep{czarnecki2017sobolev}. GalerkinFlow uses the simpler finite-difference analogue: adjacent output differences are compared with those of the paired fine target. This term is computed entirely from observed samples. It encourages local variation shared by image edges and scientific fields, but it is neither a dimensionally scaled physical-gradient norm nor related to a governing equation.

\section{Method}
\subsection{Problem Setting}
Let $\mathbf{x}_{\mathrm{lr}}\in\mathbf{R}^{C\times h\times w}$ be a $C$-channel downsampled observation, let $s>1$ be the requested scale, and let $\mathbf{y}\in\mathbf{R}^{C\times H\times W}$ be the paired fine target. The target dimensions are determined by $s$ and the input grid. Super-resolution learns
\begin{equation}
\begin{array}{rcl}
 {\cal G}_{\theta}:
 \mathbf{R}^{C\times h\times w}\times{\cal S}
 &\longrightarrow&
 \mathbf{R}^{C\times H\times W}\\
 (\mathbf{x}_{\mathrm{lr}},s)
 &\longmapsto&
 \widehat{\mathbf{y}}
\end{array}
\label{eq:operator}
\end{equation}
Here $\theta$ denotes all trainable parameters and ${\cal S}$ is either one fixed scale or a set of training scales. No differential operator, physical coefficient, forcing, boundary condition, or temporal derivative is supplied to ${\cal G}_{\theta}$ or its loss.

The downsampled observation and target lie on different grids. We apply a fixed lifting operator ${\cal U}_s$ to place the observation on the target grid,
\begin{equation}
    \mathbf{x}_0={\cal U}_s(\mathbf{x}_{\mathrm{lr}}),
    \qquad
    \mathbf{d}_s=\mathbf{y}-\mathbf{x}_0,
    \qquad
    \mathbf{x}_0\in\mathbf{R}^{C\times H\times W}
    \label{eq:coarse_anchor}
\end{equation}
where $\mathbf{x}_0$ is the coarse anchor and $\mathbf{d}_s$ is the paired coarse-to-fine displacement. The experiments use bicubic lifting, so it has no trainable parameters.

The condition $\mathbf{c}_s$ denotes the lifted coarse anchor together with four broadcast maps: vertical and horizontal scale values and target-cell sizes $1/H$ and $1/W$. In the network, the coarse anchor is first encoded by an RDN feature extractor, while the scale-cell maps are concatenated directly. Fully convolutional features and Galerkin aggregation keep the learned parameter shapes independent of $H$ and $W$. This gives architectural support for arbitrary uniform output grids. Generalization to an unseen scale is still an empirical property and depends on which scales are represented during training.

\subsection{Model Architecture}
\begin{figure*}[t]
    \centering
    \includegraphics[width=0.8\textwidth]{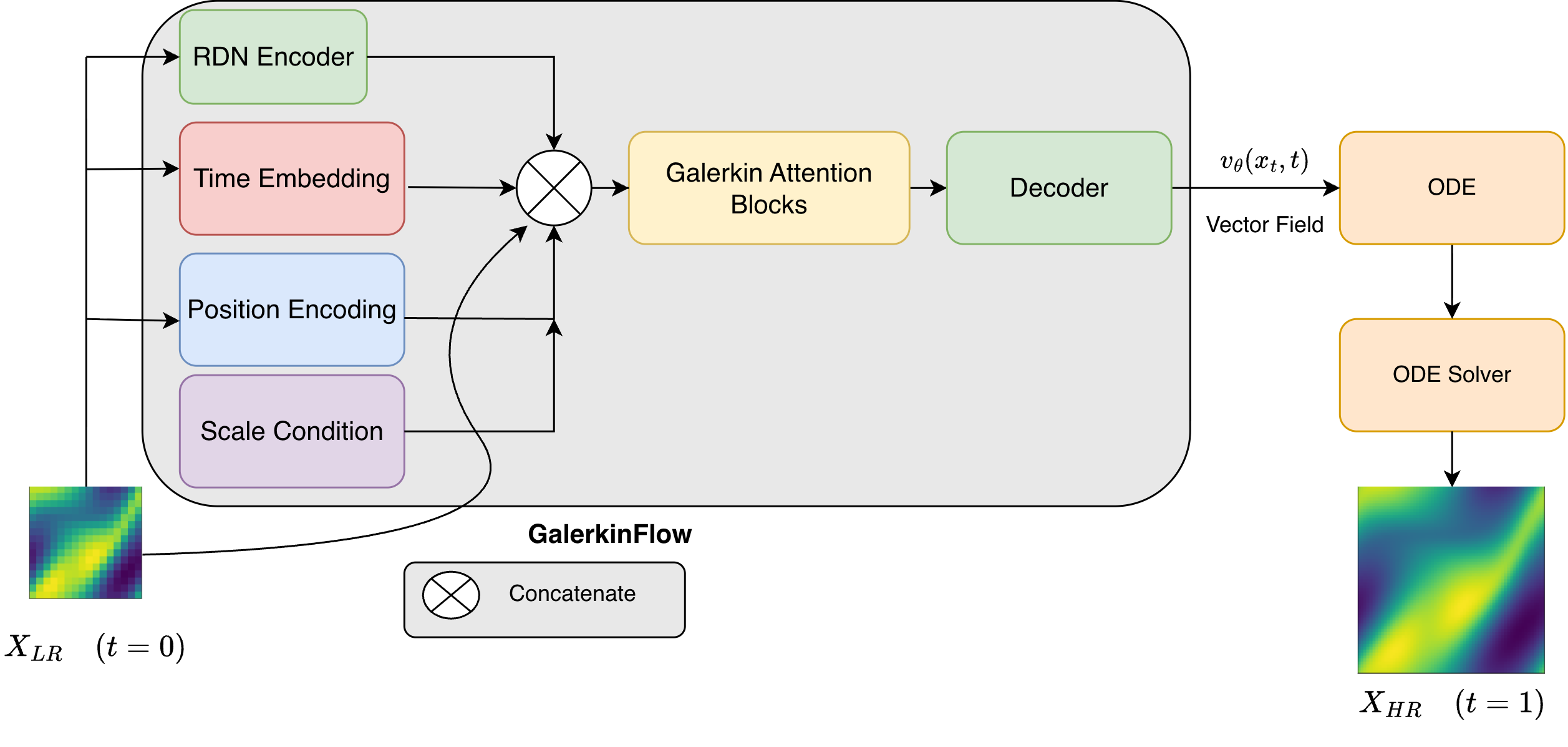}
    \caption{
    The Architecture of GalerkinFlow model.
    }
    \label{fig:galerkinflow}
\end{figure*}
\begin{figure*}[t]
    \centering
    \includegraphics[width=0.8\textwidth]{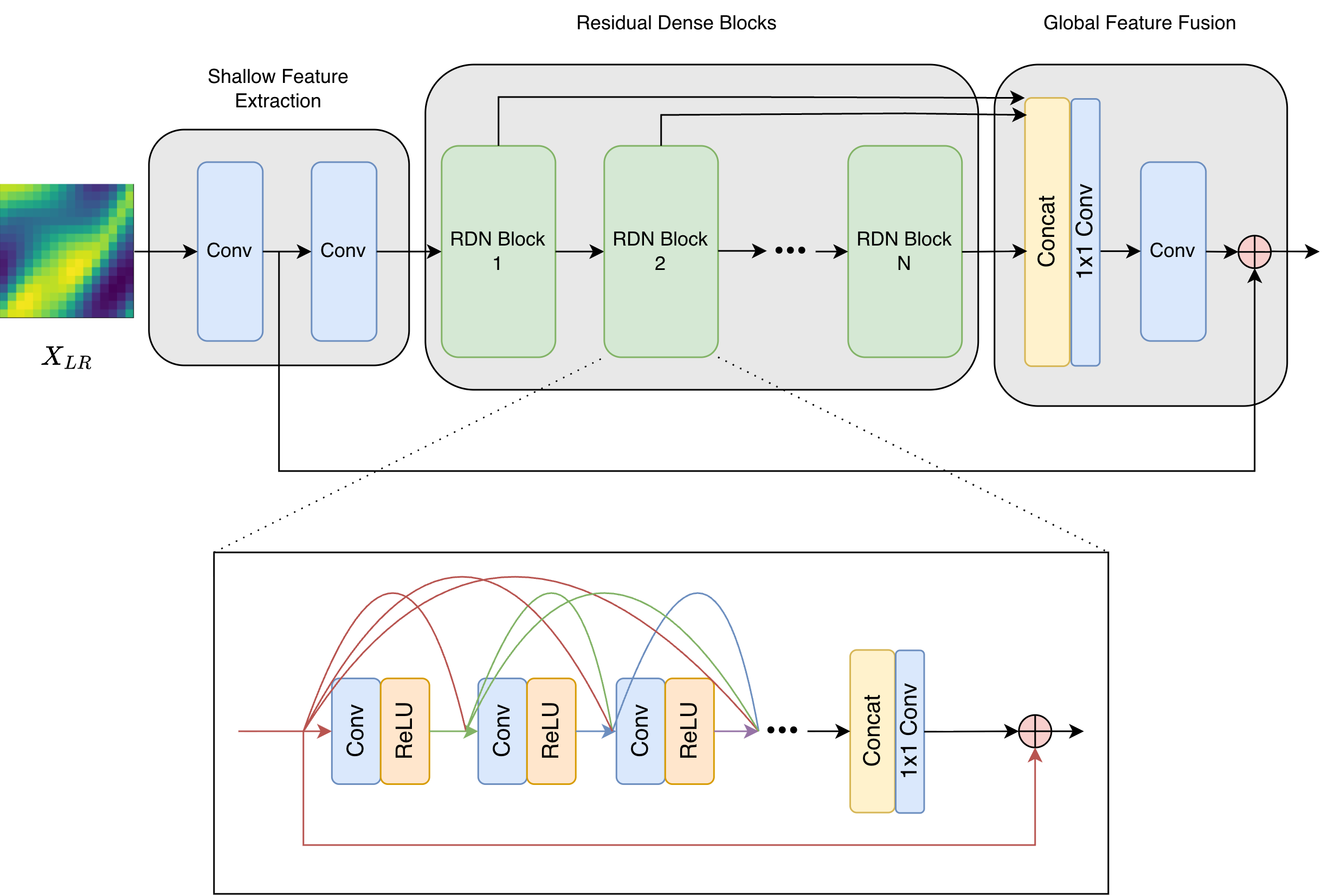}
    \caption{
    Architecture of Residual Dense Networks (RDN). There are skip connections at the end of each dense block, and convolution layers in the block are densely connected to each other.
    }
    \label{fig:rdn}
\end{figure*}

As shown in Figure \ref{fig:galerkinflow}, the GalerkinFlow combines local convolutional features with global operator mixing. An RDN encoder maps the lifted coarse anchor $\mathbf{x}_0$ to neighborhood-aware features \citep{zhang2018rdn}, as illustrated in Figure \ref{fig:rdn}. At every target-grid location, the network concatenates the current state $\mathbf{x}_t$, positional encodings, a broadcast time embedding, the encoded features, and the scale conditions. A $1\times1$ convolution lifts this tensor to $d$ hidden channels.

The lifted field passes through Galerkin attention blocks \cite{cao2021galerkin}. For one head, let $\mathbf{Q},\mathbf{K},\mathbf{V}\in\mathbf{R}^{n\times d_h}$ be query, key, and value arrays over $n=HW$ target locations. After normalizing keys and values, the global association is
\begin{equation}
    {\cal A}(\mathbf{Q},\mathbf{K},\mathbf{V})
    =\mathbf{Q}\left(\frac{\mathbf{K}^{T}\mathbf{V}}{n}\right).
    \label{eq:attention}
\end{equation}
The multiplication order avoids forming an $n\times n$ matrix. One head costs $O(nd_h^2)$ rather than $O(n^2d_h)$. On a uniform grid with locations $\{\mathbf{\xi}_i\}_{i=1}^{n}$, the central matrix has the quadrature form
\begin{equation}
 \frac{\mathbf{K}^{T}\mathbf{V}}{n}
 =\frac{1}{n}\sum_{i=1}^{n}
 \mathbf{k}(\mathbf{\xi}_i)^{T}\mathbf{v}(\mathbf{\xi}_i)
 \simeq \int_{\Omega}\mathbf{k}(\mathbf{\xi})^{T}
 \mathbf{v}(\mathbf{\xi})\,d\mu(\mathbf{\xi}).
 \label{eq:quadrature}
\end{equation}
This normalized association has been interpreted as a learned Petrov--Galerkin-type projection \citep{cao2021galerkin}. Coordinates and cell sizes preserve information about the requested discretization, while the learned attention parameters remain defined as the target grid changes. A final projection layer produces the residual field.
\subsection{Path-Conditioned Endpoint Residual Field}

Conventional endpoint regression uses $(\mathbf{x}_{\mathrm{lr}},\mathbf{y})$ only through a terminal reconstruction loss. We instead define the straight paired path
\begin{equation}
    \mathbf{x}_t=(1-t)\mathbf{x}_0+t\mathbf{y}
    =\mathbf{x}_0+t\mathbf{d}_s,
    \qquad t\in[0,1]
    \label{eq:path}
\end{equation}
The parameter $t$ denotes restoration progress. The implementation supervises the residual vector field from each sampled state to the fine endpoint,
\begin{equation}
    \mathbf{r}_t=\mathbf{y}-\mathbf{x}_t
    =(1-t)\mathbf{d}_s
    \label{eq:residual_target}
\end{equation}
Thus the learned vector field is endpoint-directed rather than the constant derivative $d\mathbf{x}_t/dt=\mathbf{d}_s$ of the linear path:
\begin{equation}
    \mathbf{v}_{\theta}(\mathbf{x}_t,t,\mathbf{c}_s)
    \simeq \mathbf{r}_t
    \label{eq:vectorfield}
\end{equation}
At training time, $t$ is sampled uniformly from $[0,1)$, so each pair supplies residual-field constraints at many partially restored states. This does not create independent labels; it reuses the same observed pair while changing the state at which the predictor is supervised.

This target has a direct one-step reconstruction interpretation. From a sampled state, define
\begin{equation}
    \widetilde{\mathbf{y}}_t
    =\mathbf{x}_t+\mathbf{v}_{\theta}(\mathbf{x}_t,t,\mathbf{c}_s)
    \label{eq:pseudoendpoint}
\end{equation}
Because $\mathbf{y}=\mathbf{x}_t+\mathbf{r}_t$,
\begin{equation}
\begin{array}{rcl}
 \|\widetilde{\mathbf{y}}_t-\mathbf{y}\|_1
 &=&\|\mathbf{v}_{\theta}-\mathbf{r}_t\|_1\\
 \|\widetilde{\mathbf{y}}_t-\mathbf{y}\|_2^2
 &=&\|\mathbf{v}_{\theta}-\mathbf{r}_t\|_2^2
\end{array}
\label{eq:equivalence}
\end{equation}
Residual-field regression therefore controls the endpoint predicted from every sampled state. This relation motivates path supervision without claiming that the constructed segment is a physical evolution.

The implemented residual field loss is
\begin{equation}
\begin{array}{rcl}
 {\cal L}_{\mathrm{vf}}
 &=&\lambda_1\,{\mathbb{E}}\left[
 D^{-1}\|\mathbf{v}_{\theta}-\mathbf{r}_t\|_1\right]\\
 &&+\lambda_2\,{\mathbb{E}}\left[
 D^{-1}\|\mathbf{v}_{\theta}-\mathbf{r}_t\|_2^2\right]
\end{array}
\label{eq:vfrisk}
\end{equation}
where $\lambda_1$ and $\lambda_2$ denote the coefficients, $D=CHW$ and the arguments of $\mathbf{v}_{\theta}$ follow Equation~\ref{eq:vectorfield}. 
\subsection{ODE Endpoint Reconstruction Loss}
The residual-field loss supervises the direction from sampled intermediate states to the paired endpoint. At test time, however, the target endpoint is unknown and the model must recover it only from the coarse anchor. The field loss in Equation~\ref{eq:vfrisk} provides local residual supervision along the constructed path, but it does not directly penalize the endpoint produced by this inference procedure. To align training with inference, we run the same endpoint solver during training and compare its output with the paired target.
\begin{equation}
    \widehat{\mathbf{y}}
    =\mathbf{x}_0+
    \mathbf{v}_{\theta}(\mathbf{x}_0,0,\mathbf{c}_s).
    \label{eq:onestep}
\end{equation}
The implementation evaluates Equation~\ref{eq:onestep} with explicit Euler updates
\begin{equation}
    \mathbf{x}^{(k+1)}=\mathbf{x}^{(k)}+
    \Delta t\,\mathbf{v}_{\theta}
    (\mathbf{x}^{(k)},t_k,\mathbf{c}_s).
    \label{eq:euler}
\end{equation}
The reconstruction term trains this predicted endpoint with
\begin{equation}
    {\cal L}_{\mathrm{rec}}
    =D^{-1}\|\widehat{\mathbf{y}}-\mathbf{y}\|_1.
    \label{eq:recloss}
\end{equation}
\subsection{Endpoint Gradient Consistency Loss}

The reconstruction loss aligns the endpoint values produced by the ODE solver with the paired target. For super-resolution, value agreement alone does not explicitly constrain how neighboring cells or pixels change relative to one another. This matters for image edges and textures, and also for sharp spatial variations in PDE fields. We therefore add a second endpoint-level penalty that compares first-order finite differences of the predicted and target endpoints.

Let $\mathbf{e}=\widehat{\mathbf{y}}-\mathbf{y}$, and let $\Delta_x$ and $\Delta_y$ denote adjacent horizontal and vertical differences. The endpoint variation term is
\begin{equation}
    {\cal L}_{\mathrm{grad}}
    =D_x^{-1}\|\Delta_x\mathbf{e}\|_1
    +D_y^{-1}\|\Delta_y\mathbf{e}\|_1,
    \label{eq:gradloss}
\end{equation}
where $D_x$ and $D_y$ are the corresponding numbers of differences. Pointwise reconstruction controls values, while this term penalizes disagreement between neighboring changes. It follows the motivation of derivative-aware and Sobolev training \citep{czarnecki2017sobolev}, but it is computed on the reconstructed endpoint, does not divide by physical grid spacing, and is not a PDE residual.

\subsection{Full Training Objective}
The complete training objective is
\begin{equation}
    {\cal L}={\cal L}_{\mathrm{vf}}
    +\lambda_{rec}{\cal L}_{\mathrm{rec}}
    +\lambda_{grad}{\cal L}_{\mathrm{grad}}.
    \label{eq:loss}
\end{equation}
where $\lambda_{rec}$ and $\lambda_{grad}$ denote the coefficients for the reconstruction and gradient losses.

The source, target, and intermediate states are paired and deterministic. GalerkinFlow is therefore a reconstruction model rather than a calibrated conditional generator, and its path is not a trajectory of the PDE that produced the field.

\section{Experimental Setup}

\subsection{Datasets and Protocol}

The controlled PDE benchmark uses fixed-scale specialists at $2\times$ and $4\times$ for every learned method. This separates reconstruction accuracy from unseen-scale generalization. The architecture can also be trained across scales, but an arbitrary-scale claim requires a separate protocol with held-out integer and non-integer scales. All experiments are run on Nvidia GH200 GPUs.

\textbf{Navier--Stokes \cite{navier_stokes_data}}
The scalar field is stored as trajectories of $64\times64$ frames. Training draws frames from the first 4,000 trajectories. Evaluation uses 1,000 frames from 1,000 held-out trajectories. For $2\times$, each fine frame is downsampled to $32\times32$; for $4\times$, it is downsampled to $16\times16$. The coarse field is lifted back to the target grid by the same bicubic operator supplied to every method.

\textbf{Darcy Flow \cite{darcy_flow_data}}
The target scalar field comes from the PDEBench dataset collection and has a non-square $64\times128$ grid. We use the provided train and test splits and evaluate all 1,000 test samples. Coarse observations have resolution $32\times64$ at $2\times$ and $16\times32$ at $4\times$.

\textbf{RGB images.}
The in-domain image benchmark uses the DIV2K validation images \cite{agustsson2017div2k}. To test transfer of DIV2K-trained checkpoints, we also evaluate Set5 \cite{bevilacqua2012low}, Set14 \cite{zeyde2010single}, BSD100 \cite{martin2001database}, and Urban100 \cite{huang2015single}. The benchmark uses one crop of $128\times128$ RGB pixels per image. The low-resolution input is generated by bicubic downsampling to $64\times64$ at $2\times$ or $32\times32$ at $4\times$, and predictions are compared on the original $128\times128$ crop.

\subsection{Baselines and Implementation}
For PDE fields, we compare with bicubic interpolation, U-Net \cite{ronneberger2015unet}, FNO \cite{li2021fno}, U-NO \cite{rahman2023uno}, and SRNO \cite{wei2023srno}. Each learned PDE method has a separately trained checkpoint for the same dataset and scale. The evaluation harness supplies identical raw input--target pairs and maps every prediction back from its training normalization before metrics are accumulated.

For RGB images, all learned checkpoints were trained on DIV2K. The baselines are bicubic interpolation, LIIF \cite{chen2021liif}, SRNO \cite{wei2023srno}, RDN \cite{zhang2018rdn}, and SwinIR \cite{liang2021swinir}.

The evaluated GalerkinFlow PDE checkpoints use 4 Galerkin blocks, 8 heads, and 128 hidden-channel width. The RGB checkpoints use 256 hidden-channel width and 16 heads. Training samples one $t$ per example and uses the loss weights in Equation~\ref{eq:loss}.
\subsection{Metrics}

For $N$ samples, raw-space relative $L_2$ is the mean of per-sample ratios
\begin{equation}
    \mathrm{Rel}\mbox{-}L_2=
    \frac{1}{N}\sum_{i=1}^{N}
    \frac{\|\widehat{\mathbf{y}}_i-\mathbf{y}_i\|_2}
    {\|\mathbf{y}_i\|_2}.
    \label{eq:rell2}
\end{equation}
MSE and MAE are pooled over raw scalar values. Relative $L_2$ is the primary metric because it normalizes field magnitude; MSE and MAE expose squared and absolute pointwise behavior. For RGB images, PSNR is averaged over the cropped RGB patches, SSIM is computed with data range 1 \cite{wang2004ssim}, and LPIPS uses VGG features with normalized RGB inputs \cite{zhang2018lpips}. Higher PSNR and SSIM are better; lower LPIPS is better.

\section{Results and Analysis}

\subsection{PDE Super-Resolution}

\begin{table*}[t]
\centering
\small
\setlength{\tabcolsep}{2pt}
\begin{tabular}{@{}llrrr@{}}
\toprule
\multicolumn{5}{c}{\textbf{Navier--Stokes}} \\
Scale & Method & Rel-$L_2$ & MSE & MAE \\
\midrule
$2\times$ & Bicubic  & 0.015335 & $1.777{\times}10^{-4}$ & $9.671{\times}10^{-3}$  \\
$2\times$ & U-Net  & 0.024337 & $3.831{\times}10^{-4}$ & $1.208{\times}10^{-2}$  \\
$2\times$ & FNO  & 0.009255 & $5.817{\times}10^{-5}$ & $5.809{\times}10^{-3}$  \\
$2\times$ & U-NO  & 0.104592 & $8.351{\times}10^{-3}$ & $6.811{\times}10^{-2}$  \\
$2\times$ & SRNO  & 0.006291 & $3.155{\times}10^{-5}$ & $1.795{\times}10^{-3}$  \\
$2\times$ & GalerkinFlow (ours)  & \textbf{0.000337} & \textbf{$8.094{\times}10^{-8}$} & \textbf{$1.731{\times}10^{-4}$}  \\
\midrule
$4\times$ & Bicubic  & 0.036574 & $1.023{\times}10^{-3}$ & $1.977{\times}10^{-2}$  \\
$4\times$ & U-Net  & 0.026176 & $4.520{\times}10^{-4}$ & $1.339{\times}10^{-2}$  \\
$4\times$ & FNO  & 0.030712 & $6.692{\times}10^{-4}$ & $1.897{\times}10^{-2}$  \\
$4\times$ & U-NO  & 0.295490 & $6.066{\times}10^{-2}$ & $1.946{\times}10^{-1}$  \\
$4\times$ & SRNO & 0.021681 & $3.785{\times}10^{-4}$ & $6.709{\times}10^{-3}$  \\
$4\times$ & GalerkinFlow (ours)  & \textbf{0.000365} & \textbf{$1.404{\times}10^{-7}$} & \textbf{$6.321{\times}10^{-5}$}  \\
\bottomrule
\end{tabular}
\hfill
\begin{tabular}{@{}llrrr@{}}
\toprule
\multicolumn{5}{c}{\textbf{Darcy Flow}} \\
Scale & Method & Rel-$L_2$ & MSE & MAE \\
\midrule
$2\times$ & Bicubic  & 0.011960 & $3.663{\times}10^{-5}$ & $3.709{\times}10^{-3}$  \\
$2\times$ & U-Net  & 0.013411 & $7.624{\times}10^{-5}$ & $4.316{\times}10^{-3}$  \\
$2\times$ & FNO  & 0.009905 & $2.727{\times}10^{-5}$ & $3.372{\times}10^{-3}$  \\
$2\times$ & U-NO  & 0.061559 & $9.721{\times}10^{-4}$ & $2.039{\times}10^{-2}$  \\
$2\times$ & SRNO  & 0.004657 & $6.194{\times}10^{-6}$ & $7.499{\times}10^{-4}$  \\
$2\times$ & GalerkinFlow (ours)  & \textbf{0.001010} & \textbf{$8.943{\times}10^{-7}$} & \textbf{$2.271{\times}10^{-4}$}  \\
\midrule
$4\times$ & Bicubic  & 0.036394 & $3.354{\times}10^{-4}$ & $9.192{\times}10^{-3}$  \\
$4\times$ & U-Net  & 0.020486 & $1.356{\times}10^{-4}$ & $6.359{\times}10^{-3}$  \\
$4\times$ & FNO  & 0.023478 & $1.398{\times}10^{-4}$ & $7.732{\times}10^{-3}$  \\
$4\times$ & U-NO  & 0.110599 & $3.137{\times}10^{-3}$ & $3.994{\times}10^{-2}$  \\
$4\times$ & SRNO & 0.012141 & $3.889{\times}10^{-5}$ & $2.434{\times}10^{-3}$  \\
$4\times$ & GalerkinFlow (ours)  & \textbf{0.001949} & \textbf{$1.851{\times}10^{-6}$} & \textbf{$1.884{\times}10^{-4}$}  \\
\bottomrule
\end{tabular}
\caption{PDE super-resolution at $2\times$ and $4\times$.}
\label{tab:pde_main}
\end{table*}

Table~\ref{tab:pde_main} shows the strongest evidence for GalerkinFlow. It obtains the lowest raw relative $L_2$, MSE, and MAE on every PDE dataset and scale. Relative to SRNO, the relative-$L_2$ reductions are 94.6\% on Navier--Stokes $2\times$, 98.3\% on Navier--Stokes $4\times$, 78.3\% on Darcy Flow $2\times$, and 83.9\% on Darcy Flow $4\times$. The same ordering under squared and absolute pointwise errors indicates that the improvement is not an artifact of one metric.

The scale effect differs by baseline. Bicubic and SRNO degrade substantially from $2\times$ to $4\times$, FNO improves over bicubic but trails SRNO on all four PDE settings, and U-NO is unstable in these checkpoint evaluations. U-Net improves over bicubic at $4\times$ but remains far from GalerkinFlow. GalerkinFlow keeps very small errors at both scales, which is consistent with the coarse-anchored path objective: the model is trained on residual targets from sampled restoration states while also being constrained at the actual coarse endpoint used during one-step inference.

\subsection{RGB Super-Resolution}

\begin{table}[t]
\centering
\small
\setlength{\tabcolsep}{4pt}
\begin{tabular}{@{}llrrrrr@{}}
\toprule
Scale & Method & PSNR $\uparrow$ & SSIM $\uparrow$ & LPIPS $\downarrow$  \\
\midrule
$2\times$ & Bicubic  & 29.86 & 0.8586 & 0.1885  \\
$2\times$ & SRNO  & 34.10 & 0.9300 & 0.1130  \\
$2\times$ & LIIF  & 33.89 & 0.9279 & 0.1157  \\
$2\times$ & SwinIR  & 34.16 & 0.9313 & \textbf{0.1118}  \\
$2\times$ & RDN  & 33.72 & 0.9264 & 0.1177  \\
$2\times$ & GalerkinFlow (ours)  & \textbf{38.19} & \textbf{0.9410} & 0.1271  \\
\midrule
$4\times$ & Bicubic  & 25.73 & 0.6748 & 0.3784  \\
$4\times$ & SRNO  & 28.39 & 0.7810 & 0.2932  \\
$4\times$ & LIIF  & 28.22 & 0.7763 & 0.2984  \\
$4\times$ & SwinIR  & 28.43 & 0.7825 & 0.2915  \\
$4\times$ & RDN  & 28.12 & 0.7723 & 0.3022  \\
$4\times$ & GalerkinFlow (ours) & \textbf{33.77} & \textbf{0.8228} & \textbf{0.2867}  \\
\bottomrule
\end{tabular}
\caption{DIV2K RGB super-resolution.}
\label{tab:div2k}
\end{table}

The DIV2K results in Table~\ref{tab:div2k} show that the GalerkinFlow are strong on distortion-oriented RGB metrics. GalerkinFlow obtains the highest PSNR and SSIM at both scales, improving over the best non-GalerkinFlow PSNR by 4.03 dB at $2\times$ and 5.34 dB at $4\times$. The perceptual metric gives a more mixed view: GalerkinFlow has the lowest LPIPS at $4\times$, but SwinIR remains better at $2\times$. Thus the RGB evidence is not only a transfer sanity check, but also supports competitive natural-image reconstruction under this protocol.

\begin{table*}[t]
\centering
\small
\setlength{\tabcolsep}{3pt}
\begin{tabular}{@{}llcccc@{}}
\toprule
Scale & Method & Set5 & Set14 & BSD100 & Urban100 \\
\midrule
$2\times$ & Bicubic & 33.27/0.910/0.135 & 28.39/0.842/0.182 & 27.09/0.813/0.208 & 25.03/0.779/0.206 \\
$2\times$ & SRNO & 38.20/0.943/0.079 & 31.91/0.903/0.120 & 30.18/0.889/0.146 & 31.31/0.913/0.101 \\
$2\times$ & LIIF & 38.00/0.942/0.080 & 31.75/0.900/0.122 & 30.09/0.887/0.147 & 30.93/0.906/0.106 \\
$2\times$ & SwinIR & 38.22/0.943/\textbf{0.078} & 31.92/0.903/\textbf{0.118} & 30.20/0.889/\textbf{0.146} & 31.49/\textbf{0.913}/\textbf{0.100} \\
$2\times$ & RDN & 38.06/0.943/0.081 & 31.64/0.899/0.124 & 30.03/0.886/0.149 & 30.55/0.900/0.110 \\
$2\times$ & GalerkinFlow (ours) & \textbf{38.49}/\textbf{0.967}/0.084 & \textbf{32.87}/\textbf{0.904}/0.120 & \textbf{31.44}/\textbf{0.890}/0.155 & \textbf{31.91}/0.909/0.121 \\
\midrule
$4\times$ & Bicubic & 28.28/0.797/0.261 & 24.37/0.656/0.343 & 23.47/0.603/0.401 & 21.52/0.558/0.405 \\
$4\times$ & SRNO & 33.26/0.888/0.189 & 27.31/0.765/0.265 & 25.48/0.702/0.338 & 25.19/0.729/0.276 \\
$4\times$ & LIIF & 33.16/0.886/0.190 & 27.23/0.763/0.268 & 25.40/0.698/0.341 & 25.01/0.721/0.284 \\
$4\times$ & SwinIR & 33.28/0.888/0.192 & 27.36/\textbf{0.767}/\textbf{0.264} & 25.48/0.702/\textbf{0.337} & 25.29/0.735/\textbf{0.269} \\
$4\times$ & RDN & 33.02/0.885/0.192 & 27.22/0.762/0.270 & 25.33/0.694/0.345 & 24.69/0.709/0.293 \\
$4\times$ & GalerkinFlow (ours) & \textbf{33.55}/\textbf{0.913}/\textbf{0.189} & \textbf{28.12}/0.755/0.283 & \textbf{27.03}/\textbf{0.707}/0.349 & \textbf{27.23}/\textbf{0.755}/0.295 \\
\bottomrule
\end{tabular}
\caption{Validation dataset of DIV2K-trained RGB checkpoints to Set5, Set14, BSD100, and Urban100. Each entry is PSNR/SSIM/LPIPS. Higher PSNR/SSIM and lower LPIPS are better.}
\label{tab:generalization}
\end{table*}

Table~\ref{tab:generalization} shows that the DIV2K-trained GalerkinFlow checkpoints also transfer well to the four external image datasets. GalerkinFlow has the highest PSNR in every dataset-scale setting, with gains over the best non-GalerkinFlow baseline ranging from 0.27 dB on Set5 to 1.94 dB on Urban100 $4\times$. SSIM is also strongest in most settings, except Set14 $4\times$ and Urban100 $2\times$. LPIPS is less uniformly favorable. GalerkinFlow is best on Set5 $4\times$, but image-specialized baselines especially SwinIR, remain stronger on several validation cases.

\section{Ablation Study}
\begin{table}[t]
\centering
\footnotesize
\setlength{\tabcolsep}{1.25pt}
\begin{tabular}{@{}lrrr@{}}
\toprule
Losses & Rel-$L_2$ raw & MSE raw & MAE raw \\
\midrule
Vector field + rec + grad & \textbf{$3.369{\times}10^{-4}$} & \textbf{$8.094{\times}10^{-8}$} & \textbf{$1.731{\times}10^{-4}$} \\
Vector field + rec & $1.068{\times}10^{-3}$ & $7.255{\times}10^{-7}$ & $4.374{\times}10^{-4}$ \\
Vector field only & $2.215{\times}10^{-3}$ & $3.407{\times}10^{-6}$ & $9.129{\times}10^{-4}$ \\
\bottomrule
\end{tabular}
\caption{Loss ablation on Navier--Stokes $2\times$.}
\label{tab:ablation}
\end{table}

Table~\ref{tab:ablation} shows a consistent improvement as the two endpoint-level objectives are added on Navier--Stokes $2\times$. Relative to vector-field supervision alone, adding endpoint reconstruction reduces Rel-$L_2$, MSE, and MAE by 51.8\%, 78.7\%, and 52.1\%, respectively. This supports the role of ${\cal L}_{\mathrm{rec}}$ in anchoring training to the coarse state encountered by the ODE solver at inference. Adding gradient consistency yields a further reduction of 68.5\%, 88.8\%, and 60.4\% over the no-gradient variant, giving the full objective the lowest error under all three metrics. Thus, in this setting, residual-field supervision establishes the restoration direction, endpoint reconstruction constrains the integrated solution, and gradient consistency provides complementary control of local spatial variation. 

All tables report one-step reconstruction from the lifted coarse anchor. Intermediate path states enlarge the supervised region during training, but they are not additional inference iterations in these results. The same learned field can be evaluated with more Euler intervals, yet that changes the compute budget and should be considered as an accuracy--cost tradeoff.

The PDE and RGB outcomes also clarify where the current backbone matters. On PDE fields, the CNN--Galerkin residual-field model and coarse-anchored objective are evaluated in the same scalar raw space used to train the checkpoints, and the gains are large across every error metric. On RGB images, the same design now gives strong PSNR and SSIM, but LPIPS shows that perceptual-feature alignment remains less consistently improved than pointwise fidelity. This split motivates future work on lighter image-specialized residual-field models, explicit perceptual objectives, and parameter-matched PDE retraining before attributing all gains to any single architectural component.

\section{Conclusion}

This work investigated how much supervision can be extracted from a paired coarse--fine sample beyond conventional endpoint regression. GalerkinFlow interprets each pair as a deterministic restoration path and trains a scale-conditioned residual field at sampled intermediate states, where the exact target is the remaining displacement to the fine endpoint. The pathwise objective is complemented by ODE endpoint reconstruction from the coarse anchor actually encountered at inference and by finite-difference consistency at the reconstructed endpoint. Together, these terms connect intermediate-state supervision, final-value accuracy, and local spatial variation without requiring a governing equation, physical coefficients, or solver metadata. A CNN encoder and Galerkin operator backbone provide the local feature extraction and global mixing needed to realize this objective on both scientific fields and RGB images.

Fresh checkpoint evaluations establish the strongest evidence on scientific super-resolution. GalerkinFlow achieves the lowest Rel-$L_2$, MSE, and MAE among the evaluated equation-agnostic baselines on both Navier--Stokes and Darcy Flow at $2\times$ and $4\times$. The Navier--Stokes $2\times$ ablation further shows a consistent progression: endpoint reconstruction improves over vector-field supervision alone, and gradient consistency improves all three errors again. GalerkinFlow also obtains the highest PSNR and SSIM on DIV2K at both scales and the highest PSNR across all reported Set5, Set14, BSD100, and Urban100 transfer settings. LPIPS is less uniform, with image-specialized models retaining an advantage in several cases, so the evidence is strongest for distortion-oriented fidelity rather than perceptual-feature alignment.

Taken together, these results support a supervision principle: a paired endpoint does not provide only one training constraint, but determines a continuum of endpoint-directed residual targets along a task-relevant path. Exploiting this structure can constrain how reconstruction behaves before reaching the endpoint while the explicit coarse-anchor loss preserves alignment with one-step inference. The contribution is therefore neither a new physical solver nor a conditional generative model; it is an equation-agnostic way to organize supervision already contained in paired data. This perspective is especially useful when only low-resolution observations are available and the equations or acquisition process are unknown.

Several limitations define the next steps. The current benchmarks use dataset-specific checkpoints, and most use scale-specific checkpoints, so they do not show arbitrary-scale reconstruction with one shared model. Though GalerkinFLow can query one checkpoint on arbitrary uniform target scales, the accuracy is decreased. The loss ablation covers only Navier--Stokes at $2\times$, and therefore does not establish universal loss weights. Moreover, the constructed path is not a physical trajectory, the finite-difference term is not a PDE residual, and no conservation law or boundary condition is guaranteed. Future work should evaluate shared-scale checkpoints, and perceptual objectives, parameter-matched backbones, and physically constrained variants, while also quantifying the accuracy--cost tradeoff of multi-step ODE integration.

\bibliography{aaai2027}


\end{document}